# AI-Driven Innovations in Modern Cloud Computing

Animesh Kumar[1,2,*]

[1]Master of Computer Science, Illinois Institute of Technology, Illinois, USA
[2]Bachelor of Engineering in Computer Technology, Nagpur University, Maharashtra, India

**Abstract**  The world has witnessed rapid technological transformation, past couple of decades and with Advent of Cloud computing the landscape evolved exponentially leading to efficient and scalable application development. Now, the past couple of years the digital ecosystem has brought in numerous innovations with integration of Artificial Intelligence commonly known as AI. This paper explores how AI and cloud computing intersect to deliver transformative capabilities for modernizing applications by providing services and infrastructure. Harnessing the combined potential of both AI & Cloud technologies, technology providers can now exploit intelligent resource management, predictive analytics, automated deployment & scaling with enhanced security leading to offering innovative solutions to their customers. Furthermore, by leveraging such technologies of cloud & AI businesses can reap rich rewards in the form of reducing operational costs and improving service delivery. This paper further addresses challenges associated such as data privacy concerns and how it can be mitigated with robust AI governance frameworks.

## 1. Introduction

AI is one of the fundamental drivers that technology has contributed to in modern cloud computing, making it enter a new realm of innovation and efficiency. Cloud computing provides IT resources in a flexible and scalable way and with AI, cloud computing can deliver previously unseen capabilities like applications with integration of machine learning or deep learning. With the power of AI and Cloud Computing enterprises can ingest and analyse massive volumes of data quickly and with great accuracy facilitating more intelligent decision-making for better strategic decisions.

Cloud computing combined with advanced AI technologies helps businesses to automate repetitive tasks and optimize resource management which reduces operational overheads. Even with such integration applications on cloud utilize computing power only when required, alert maintenance before it is needed and be managed with next-to-no human intervention. Applications deployed in the cloud can scale infinitely based on the requests/ demand and further strengthened by AI capabilities. Few use cases where combination of cloud & AI are real-time threat and anomaly detection, to cloud security systems protect against harmful cyber threats.

Furthermore, AI also enhances the user experience with demanding personal recommendations and more advanced natural language processing. AI-influenced virtual assistants, and chatbots make the interactions intuitive; whereas predictive analytics tools predict future trends or customer requirements for better insights. With further developments on the horizon for these technologies, their convergence stands to change cloud computing as we know it–which will open up whole new avenues of growth and innovation in multiple industries. Harnessing remote computation power over the Internet without the need for expensive hardware and making costly services available to mass users at a marginal cost gave birth to the concept of cloud computing, [1].

* Corresponding author:
animesh21@gmail.com (Animesh Kumar)

## 2. Literature Review

| Topic | Authors and Year | Key Findings | Source |
|---|---|---|---|
| Enhanced Data Analytics | Xu et al. (2022) | AI-driven analytics platforms enable efficient data processing and predictive analytics, improving decision-making. | *Journal of Computing Research* |
| | Zhang & Wang (2023) | AI tools enhance real-time data analysis capabilities, crucial for timely insights in various sectors. | *International Journal of Cloud Computing* |
| Automation and Operational Efficiency | Liu et al. (2021) | AI-driven automation reduces manual intervention in cloud management, optimizing resource management and reducing costs. | *Cloud Computing Journal* |
| | Singh & Sharma (2022) | Automated cloud services improve system reliability and reduce downtime through predictive maintenance. | *IT Operations Journal* |
| Advanced Security Measures | Gupta & Sinha (2023) | AI enhances threat detection and response by identifying anomalies and predicting security incidents. | *Cybersecurity Review* |
| | Patel et al. (2022) | AI-driven security measures provide robust defense mechanisms against sophisticated cyber-attacks. | *Journal of Information Security* |
| Personalized User Experiences | Chen & Lee (2023) | AI algorithms deliver personalized content and interactions, enhancing user engagement and satisfaction. | *International Journal of Cloud Computing* |
| Emerging Trends and Future Directions | Smith & Johnson (2024) | Integration of edge computing with AI enables real-time processing and decision-making close to data sources. | *Journal of Computing Research* |
| | Brown et al. (2024) | Convergence of AI with quantum computing could enhance computational capabilities and problem-solving. | *Journal of Computing Research* |
| | White & Patel (2023) | Emphasis on ethical considerations and governance frameworks to address privacy, bias, and transparency in AI applications. | *Ethics in Technology Review* |

## 3. What is Modern Cloud Computing

Cloud computing today is providing on-demand computing services making it infinitely scalable, flexible without involving time consuming process of procurement and deployment. In recent years, the landscape of computing paradigms has witnessed a gradual yet remarkable shift from monolithic computing to distributed and decentralized paradigms such as Internet of Things (IoT), Edge, Fog, Cloud, and Serverless. The frontiers of these computing technologies have been boosted by shift from manually encoded algorithms to Artificial Intelligence (AI)-driven autonomous systems for optimum and reliable management of distributed computing resources, [2]. This is a radical departure from legacy IT infrastructure and it comes with a number of important capabilities as well:

*Some of the Main Characteristics Of New Era Cloud computing*

**On-Demand Self-Service:** Businesses want to use computing resources like server time and network storage, without requiring human intervention from the service provider. It provides dynamic scaling as per the real-time requirements.All that is required is a laptop with an internet connection. Armbrust et al. (2010), Cloud computing, the long-held dream of computing as a utility, has the potential to transform a large part of the IT industry, making software even more attractive as a service and shaping the way IT hardware is designed and purchased, [13].

**Resource Pooling**: the cloud providers pool computing resources to serve multiple customers using a multi-tenant model. They offer business options to share storage, processing power and memory resources among multiple users which ideally improves utilization efficiency and is operationally highly economical.

**Rapid Elasticity:** Ability to quickly expand and contract resources as the requirement in real-time. This flexibility allows businesses to efficiently manage fluctuating workloads, from traffic peaks through ongoing growth.

**Measured Service:** cloud computing resources charges are based on the quantity of services and utilization of such accessed. Cloud offers different types of pricing such as the pay-as-you-go model, reserved instance model, dedicated instances model making it easier for organizations to control their expenses and access what they need, without any upfront investment or long-term commitment.

Last decade witnessed a huge paradigm shift in the field of computing world. This shift is from traditional stand alone systems to Cloud Computing which is a shared pool of hardware and software resources combined to provide on demand services [16,17].

**Service Models:**

*Infrastructure as a Service (IaaS):* Almost all cloud providers offer IaaS as a basic offering. The services include virtualized servers, computing, and elastic storage. For example, Amazon Web Services (AWS) and Microsoft Azure.

*Platform as a Service (PaaS):* Where providers such as Google (App Engine), Heroku extend their readymade platform for businesses allowing them to develop, run and manage applications without specifically managing the underlying infrastructure.

*Software as a Service (SaaS):* The model in which businesses offer their software to be used by consumers on subscription basis. typically a web-based software that consumers/ businesses can use without downloading or installing by paying monthly, annual or other period fees. For instance, Microsoft Office

365 and Salesforce.

Cloud computing is able to provide a variety of services at the moment but the main three services are InfrastructureAs-A-Service, Platform-As-A-Service and Software-As-AService also called as service model of Cloud computing, [14].

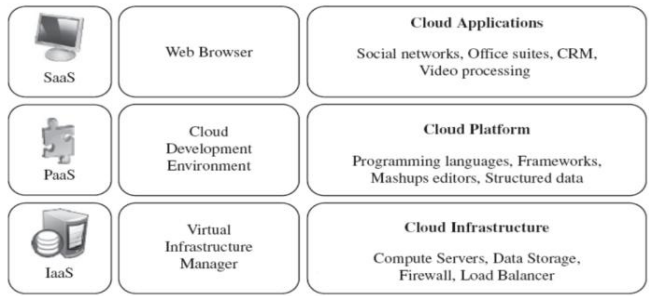

**Figure 1.**　Cloud computing service model [15]

**Deployment Models:**

With modern cloud computing various types of deployment could be achieved in line with the need, compliance and regulations.

**Public Cloud**: Services which are shared across multiple users and accessed via the public internet. Due to its nature it is highly cost-effective and scalable but offers less control over security.

**Private Cloud**: Its highly secured environment and is dedicated to a single organization offering more control over dedicated resources allocated.

**Hybrid Cloud**: Combines public and private clouds, allowing organizations to use both for different needs and integrate them for a more flexible approach.

**Security and Compliance**: Modern cloud computing comes with varied advanced security features which can be added to applications with ease. The features include data encryption, identity management and related services to ensure data protection and adherence to regulatory compliance.

Cloud implementation is the process of creating a virtual computing environment. Deployment in the cloud provides organizations with flexible and scalable virtual computing resources. A cloud deployment model is the type of architecture in which a cloud system is deployed. These models differ in terms of administration, ownership, access control, and security protocols, [18].

**Benefits of Modern Cloud Computing**

*Cost Efficiency:* Modern cloud computing extends services which reduces the capital expenditure by requiring no physical hardware and is opex paired with usage based charges.

*Flexibility and Agility:* Businesses can access computing power on-demand and also distribute applications across regions for aiding innovation & response to market changes.

*Disaster recovery and backup*: an important aspect of cloud computing is its option for disaster recovery & backup. Unlike traditional methods which needs huge infra, cloud computing offers elastic storage clubbed with economical benefits.

*Collaboration and Accessibility:* When an application is deployed on modern cloud infrastructure, It allows sharing resources as it can access different services conveniently from anywhere.

Cloud computing has shown to be advantageous to both consumers and corporations. To be more specific, the cloud has altered our way of life. Overall, cloud computing is likely to continue to play a significant role in the future of IT, enabling organizations to become more agile, efficient, and innovative in the face of rapid technological change. This is likely to drive further innovation in AI and machine learning in the coming years, [19].

## 4. How AI Impacts Modern Cloud Computing

AI is significantly shaping the landscape of contemporary cloud computing, reconfiguring the approaches to delivering and managing the services in new ways. The AI powered services improve Multiple Dimensions Of Cloud Computing which boosts the capabilities of the infrastructure and services available on the cloud, making them more efficient than before. Automating processes that earlier required manual inputs and also speeding new innovative ideas.

The diversity of large distributed application means there is a requirement for effective big data analytics mechanisms to process the required data in an efficient manner using innovative data processing techniques, [20].

*The following are few of the key ways AI influences a cloud computing approach today:*

**Enhanced Data Analytics:** Using AI-driven analytics in cloud environments has made it possible to process and analyze the vast amount of data. Machine learning algorithms can identify patterns and trends in data to help businesses find insights they never would have discovered with traditional methods allowing them to make faster decisions. Further, AI combined with Cloud computing can offer predictive analytics (predicting future events) and behavioural analysis of customers which can help businesses to drive growth.

Example: AI based tools in cloud platforms such as AWS SageMaker or Google Cloud AI could be used to analyze a wide range of data sets for further insights and forecasting which would help improve business strategies and operational efficiency.

**Automation and Efficiency:** AI allows automation of different cloud management processes, which in turn decreases the dependency on manual actions and renders seamless performance. Some of the work performed in AI-driven automation includes resource provisioning, load-balancing handling and system maintenance. This enables better usage of resources, reduced costs and simpler operations.

Example: by means of AI, services like Microsoft Azure's Automation or AWS Lambda can automate boring, repeatable tasks such as resizing resources on demand (up-or down-scaling) with no human intervention.

Artificial intelligence is said to play an important role for telecom companies as it is customer profiles and provide them with offers based on their needs and interest. Moreover, AI supports operation team, by detecting and predicting system failure, and thereby provides instant corrective action. Furthermore, AI can tackle customer service, for example, TOBi is a chatbot introduced by Vodafone to help customers by answering enquires online, support for problem trouble shooting, that will increase customer satisfaction [21,22].

**Improved Security:** The Advancements of AI powered Threat Detection and Prevention also improves cloud security with advanced threat detection and response capabilities. Network traffic analysis by machine learning models allow network administrators to be more efficient at diagnosing anomalies and detecting threat-poses. Moreover, AI-powered security solutions can evolve and counteract a new thread very quickly to enhance overall cyber-securities.

Example: Uses of AI for detecting suspicious activities and preventing a breach in the cloud security like AWS GuardDuty,

**Personalized User Experiences:** The AI algorithms in modern cloud service are used to analyze the behavior and preferences of customers, using which businesses can tailor make personalized experiences. AI powered applications on cloud providing a personalized experience increases user engagement and satisfaction driving sales growth.

Example: SaaS platforms like Salesforce leverage AI to offer personalized customer relationship management (CRM) features that deliver customized recommendations and insights based on user data.

**Smart Resource Management:** Resource Allocation And Management Using AI For Cloud Optimization Models built on such predictive analytics and related machine learning can help forecast resource demand or utilization in the cloud, identify ways to enforce optimal uses of those resources, and anticipate inefficiencies. Because lightweight resources can be suspended and resumed at will, so that saves a few dollars and optimizes the use of cloud infrastructure.

AI-driven capabilities in cloud platforms that forecast consumption behaviors and adjust allocation of resources," auto-scaling" to avoid instance overcrowding or under usage.

**Advanced Natural Language Processing (NLP):** AI-based natural language processing (NLP) tools offered in modern cloud computing improve user interactions and help facilitate customer service capabilities. With NLP, there is a far easier type of communication like chatbots and virtual assistants or direct voice commands.

Examples: Google Cloud Natural Language API and Microsoft Azure Cognitive Services are examples of NLP services to build higher level conversational agents or text analysis applications.

**Predictive Maintenance & Improvements:** The predictive maintenance feature of AI processing capabilities in cloud computing can detect potential issues before they affect services. Proactively maintaining and reducing downtime of the system, as machine learning models analyze historical data and usage patterns to predict failures.

Cloud platforms continuously check the health state of the environment and predict failure for hardware or software approaches in time to optimize performance using AI.

Another challenge is investigating AI technology to perform predictive analytics. They are using cloud services to build and train machine learning models that help them to gain valuable, actionable insights, [25].

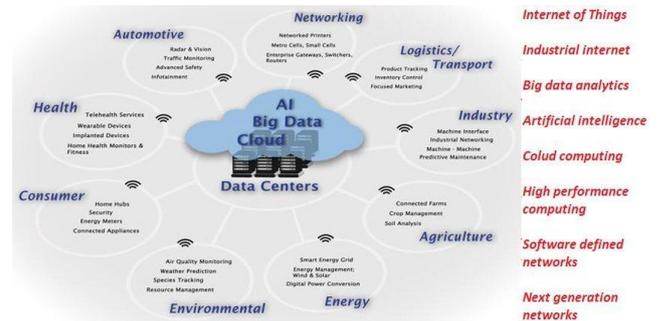

IoT, cloud computing, big data and artificial intelligence -the new drivers of the ICT ecosystem. [27].

## 5. Algorithms in Cloud Computing

With rapid technology advancements, more and more AI services are being added with services of cloud computing which would help businesses developments using those services. Few of the AI algorithms are commonly utilized, including:

1. **Machine Learning (ML)**: Algorithms such as decision trees, random forests, and neural networks enable predictive analytics and data classification.
2. **Natural Language Processing (NLP)**: Techniques like sentiment analysis and chatbots are heavily used enhance user interaction and data interpretation.
3. **Deep Learning**: Convolutional and recurrent neural networks are used for image recognition and sequence prediction, facilitating complex data analysis.
4. **Reinforcement Learning**: This approach optimizes resource allocation and automated decision-making by learning from interactions with the environment.
5. **Clustering Algorithms**: Techniques like k-means and hierarchical clustering help in segmenting data for targeted insights and analytics.

These algorithms, when deployed in cloud environments, can leverage scalability and computational power to deliver advanced AI capabilities.

## 6. Future Scope

The future of AI in cloud computing holds phenomenal potential for enhanced automation and real-time data processing. As technology evolves, the services can extend robust solutions for data security and bias mitigation, driving broader adoption and trust in AI systems.

*Here are the few areas where cloud computing with AI could potentially grow in future.*

**Edge Computing Integration:** with the rise of edge computing in mainstream application deployments, AI features will add to processing large data close to where it is generated. This capability would allow immediate decision making and eliminate latency in use cases that demand instant response, for instance heavily used in self-driving vehicles or smart cities.

**Better Personalization and UI**: The AI in the cloud platforms can extend sophisticated recommendation systems and natural language understanding, which will allow even more personalized experiences. This will drive more seamless interactions and personalized experiences through any service from e-commerce to healthcare.

**Evolution of Security:** AI will further evolve cloud computing security by creating increasingly sophisticated threat detection and response capabilities which can be integrated in next-generation solutions. Next-generation products will have more predictive security, self healing systems and adaptive defenses that ensure companies are able to anticipate new threats or stop them before they ever disrupt their business.

**Quantum Computing Incorporation**: The confluence of AI, cloud computing and quantum holds transformative capabilities. With quantum computing, the potential to solve otherwise intractable problems at speeds previously unimaginable would take AI capabilities a giant step further and enable entirely new applications for data analysis, optimization or problem solving.

**Ethical AI and Governance:** As the evolution of artificial technology continues to grow, a line will be drawn on how it can be applied regarding ethical application as well as governance. Moving forward, the technology landscape will see models of frameworks and standards for responsible AI use in the cloud: including that address concerns around privacy, bias & transparency to ensure both effectiveness and equity.

Through deep learning, machines can use existing training data in industry to analyze large amounts of data retrieved through data mining. AI will master this data such as traffic increasing the accuracy of decisions made in this industry. Using AI, the security of systems has been increased whereby attacks can be detected automatically through machine learning. This has resulted in less attacks adding value to both the users and firms. The artificial intelligence-based technologies call for alterations in several branches of law; while interface technologies show the difficulty and complexity of regulating interdisciplinary fields, [23,24,26]. The usage of AI technologies in the field of mechanical engineering has potential to revolutionize traditional design, manufacturing, and maintenance processes. With AI-powered design tools engineers now can generate optimized designs faster with greater efficiency, leading to enhanced product performance and reduced development cycles. Further, Predictive/forecasting method of AI in maintenance systems facilitate early detection of equipment failures, thereby minimizing downtime and maintenance costs, [28].

**Potential Challenges & Benefits:** There are many benefits for AI in cloud computing, but there are also some challenges and limitations which needed to be addressed. One of the most important concerns is the Data Privacy and Security. AI systems needs extreme large amounts of data to teach how to generate appropriate answers and the protection of sensitive information also becomes extremely important. The challenge is that the cloud environment is not immune to breaches and meeting the required regulations such as GDPR can cause some hiccups in the data handling practices. As well, since data is all stored in one place, any point of failure can hit the entire dataset, which is why data recovery in different cloud region is a must.

Another limitation could be that AI algorithms can be biased. As the AI models are trained based on data, if the distribution of training data is biased or do not sufficiently capture the underlying reality of what would be expected in real use, this will result in systems perpetuating these biases and may lead to unfair outcomes through decision-making. Moreover, the level of difficulty to fit AI solutions with cloud infrastructures can avert it from capture and as result businesses may face computing struggles or require large amount of resources for the integration. Furthermore, AI on cloud computing presents another layer of challenge, as it requires trained staff to operate these systems profitably, which is especially hard for small companies on tight budgets.

The future of AI-driven cloud computing has immense potential to redraw technological boundaries and unleash new windows for innovation and efficiency even as it helps us confront novel challenges on both performance evaluation fronts along with raising ethical dilemmas.

## 7. Conclusions

To conclude, AI combined with modern cloud computing results in a huge transformation in the digital landscape and provides opportunities for businesses to extend innovative services across industries. By merging the scalability and flexibility of cloud platforms with AI intelligence & automation power businesses can reach unimaginable operational efficiencies. With enhanced data analytics and automated processes, AI has made it much easier to secure networks beyond the traditional ways, providing more opportunities for growth and personalized user experiences with an advantage unlike ever before.

Now, looking into the future, cloud computing capabilities are expected to get even bigger as AI evolves more. As more and more organizations standardize their adoption of these modern advances, they will not only be positioned to navigate the challenges presented in a digital world but also leverage every possibility from all pool data systems. The adoption of AI-based cloud services will become imperative to remain competitive and survive in the constantly changing technical environment.